%% file: iclr2025.tex
\documentclass[11pt]{article} 
\usepackage[final]{acl}
\usepackage[T1]{fontenc}
\usepackage{times}
\usepackage{xspace}
\usepackage{latexsym}
\usepackage[most]{tcolorbox}
\tcbset{
  promptbox/.style={
    enhanced,
    colback=white,
    colframe=black,
    boxrule=1pt,
    sharp corners,
    left=10pt,right=10pt,top=8pt,bottom=8pt,
    fonttitle=\bfseries\large,
    colbacktitle=black,
    coltitle=white,
    attach boxed title to top center={yshift=-2pt},
    boxed title style={
      sharp corners,
      boxrule=1pt,
      colframe=black,
      interior style={fill=black},
      height=16pt,
      top=2pt,bottom=2pt,left=10pt,right=10pt,
    },
  }
}
\usepackage{enumitem}

\usepackage{ragged2e}

\usepackage{caption}
\usepackage{enumitem}   
\usepackage{graphicx} 

\usepackage{amsmath}
\usepackage{amssymb}
\usepackage{hyperref}
\usepackage{url}
\usepackage{multirow}
\usepackage{tabularx}     
\usepackage{booktabs}     
\usepackage{array}        
\usepackage{xcolor}   
\usepackage{cleveref}
\usepackage{wrapfig}

\usepackage{todonotes}

\definecolor{gold}{rgb}{0.83, 0.69, 0.22}

\NewDocumentCommand{\steeve}
{ mO{} }{\textcolor{gold}{\textsuperscript{\textit{Steeve}}\textsf{\textbf{\small[#1]}}}}
\usepackage{pifont}  
\newcommand{\cmark}{\textcolor{green!60!black}{\ding{51}}}
\newcommand{\xmark}{\textcolor{red}{\ding{55}}}
\crefformat{section}{\S#2#1#3}
\crefformat{subsection}{\S#2#1#3}
\crefformat{subsubsection}{\S#2#1#3}
\crefrangeformat{section}{\S\S#3#1#4 to~#5#2#6}
\crefmultiformat{section}{\S\S#2#1#3}{ and~#2#1#3}{, #2#1#3}{ and~#2#1#3}

\Crefformat{figure}{#2Fig.~#1#3}
\Crefmultiformat{figure}{Figs.~#2#1#3}{ and~#2#1#3}{, #2#1#3}{ and~#2#1#3}
\Crefformat{table}{#2Tab.~#1#3}
\Crefmultiformat{table}{Tabs.~#2#1#3}{ and~#2#1#3}{, #2#1#3}{ and~#2#1#3}
\Crefformat{appendix}{#2Appx.~\S#1#3}

\newcommand{\stitle}[1]{\vspace{1ex} \noindent{\bf #1.}}

\newcommand{\modelname}{\texttt{\textbf{GTA}}\xspace}

\title{\modelname: Generating Long-Horizon Tasks for Web Agents at Scale}

\author{
\textbf{Tenghao Huang}$^{1,2\dagger}$,
\textbf{Kung-Hsiang Huang}$^{2}$,
\textbf{Prafulla Kumar Choubey}$^{2}$ \\
\textbf{Yilun Zhou}$^{2}$,
\textbf{Muhao Chen}$^{3}$,
\textbf{Jonathan May}$^{1}$,
\textbf{Chien-Sheng Wu}$^{2}$ \\
$^{1}$University of Southern California,
$^{2}$Salesforce AI Research \\
$^{3}$University of California, Davis \\
{\small $\dagger$ Work done during internship at Salesforce.} \\
\texttt{tenghaoh@usc.edu}
}
\begin{document}

\maketitle

\begin{abstract}
Web agents, which couple language models with browsing and tool-use capabilities, show promise as open web assistants. Yet progress is increasingly limited by the lack of scalable, process-level supervision. Existing benchmarks are largely \emph{manually constructed}, providing only coarse start–goal annotations without intermediate trajectories, while recent automatic generation efforts remain expensive, biased, and shallow. These limitations prevent reliable training and evaluation of agents that must generalize to realistic, multi-hop, cross-page tasks.
We introduce a scalable framework \modelname that integrates crawling, retrieval-based seeding, in-context generation, and automated quality control to produce realistic tasks paired with executable trajectories. This design decouples crawling from generation for greater efficiency, grounds tasks in the site graph to enforce compositionality, and ensures dense supervision through deterministic replays and systematic validation. 
We instantiate the pipeline on over 50 websites covering e-commerce, government, forums, and news, with multilingual and multi-hop coverage. The resulting benchmark reveals a significant human–agent performance gap and enables detailed diagnostics. Our contributions are three-fold: (i) formalizing multi-hop web-agent task generation, (ii) proposing an efficient and validated pipeline for automatic data creation, and (iii) releasing a self-evolving benchmark ecosystem where end users can generate up-to-date tasks grounded in live web content \footnote{Code available at \url{https://github.com/tenghaohuang/GTA}.}.
\end{abstract}

\input{sections/intro}

\input{sections/benchmark}

\input{sections/analysis}

\input{sections/results}

\input{sections/related_works}
\input{sections/conclusion}

\bibliography{iclr2025}

\newpage
\appendix

\section{Website Coverage}
\Cref{tab:domains} shows representative websites covered in \modelname benchmark.
\input{resources/webiste_coverage_table}

\section{Prompt Details}
\Cref{fig:multihop_prompt} shows detailed prompt of multi-hop web agent task creation. \Cref{fig:ambiguity_assessment}, \Cref{fig:check_correctness}, and \Cref{fig:concatenation_assessment} show detailed prompts of the quality control pipeline. 
\input{resources/prompts/task_generation_prompt}

\input{resources/prompts/quality_control_prompts}

\section{Performance Per Website}
\Cref{fig:performance_per_website} shows the success rates of \textsc{browseruse} and \textsc{Agentoccam} on a subset of websites.

\section{Cost of Task Generation}
\label{app:cost}

\stitle{Computational Cost} An important consideration in designing \modelname is the computational and monetary cost of generating large-scale tasks. Our pipeline is deliberately lightweight: the only substantive expense lies in producing webpage descriptions for retrieval indexing. Crawling itself incurs no cost, and in-context task generation requires only a small number of LLM calls.

Concretely, generating descriptions for approximately 2{,}000 webpages and generating 100 multi-hop tasks via in-context prompting costs around \$10. Once the site graph $\mathcal{G}$ is constructed and indexed, however, the marginal cost of additional tasks is effectively negligible: the graph enables arbitrarily many tasks to be generated without re-incurring indexing costs. By contrast, prior benchmarks such as \textsc{AgentTrek} report an average cost of \$0.55 per task, as they rely on LLM-based exploration policies to traverse websites. This design makes the marginal cost of each new task significantly higher and limits scalability, whereas our graph-based approach amortizes indexing cost and supports large-scale task generation at roughly \$0.10 per task in our experiments. This low overhead is particularly advantageous for dynamic test sets, where new tasks can be generated continuously with minimal additional expense.

\stitle{Time Cost} In addition to monetary efficiency, our pipeline is designed to be lightweight in terms of time.  
The most time-consuming stage is crawling webpages and creating retrieval index.  
For a medium-scale setting of approximately 2{,}000 webpages, this process completes within 30--40 minutes on a single machine. 
Once the site graph is built, task generation itself is relatively fast, allowing us to rapidly scale the benchmark to thousands of multi-hop tasks without significant runtime bottlenecks.  
This property stands in contrast to LLM-exploration-based benchmarks, where each new task requires running an agent rollout through the website, incurring substantial time costs per task.  

\input{resources/performance_website}

\section{Multilingual Task Performances}
\Cref{fig:multilingual} shows web agents' cross-lingual performances. Both \textsc{BrowserUse} and \textsc{AgentOccam} exhibit a consistent decline in success rate when moving from English to non-English sites. The results highlight a strong bias in current web agents toward English-centric environments, while exposing severe limitations in their ability to operate on foreign-language websites.

\input{resources/multilingual_task_fig}

\section{Annotation Details}
\label{app: annotation}
We recruit 3 annotators with computer science background for our human evaluation. An example annotation interface is shown in \Cref{fig: annotation_interface}. The inter-annotator agreements (Cohen's Kappa) is 0.3, which indicates fair agreement.

\section{Implementation Details}
\label{app: implementation}

We use GPT-4o as the underlying large language model for agent exploration and GPT-5.1 for task generation. Our web interaction stack is built on Chromium, which is used to render webpages and capture visual snapshots for agent perception. This setup matches that of prior web-agent benchmarks, enabling consistent state replay and fair comparison. Each environment step corresponds to a deterministic browser action followed by a rendered page snapshot.

We inherit environment settings and site configurations from established benchmarks where applicable (e.g., using the same website instances and task distributions as WebArena and WebDS). To ensure safety and reproducibility, we restrict the agent to publicly accessible webpages and explicitly avoid login-gated content, personal accounts, or sensitive transactional flows.

Dynamic webpage elements such as pop-ups, cookie consent banners, and modal dialogs are handled using the same rules and heuristics defined in the corresponding benchmark environments. We ensures that agent behavior remains comparable to prior baselines and that observed improvements stem from reasoning and decision-making rather than environment customization.

\input{resources/annotation_inter}
\end{document}

%% file: sections/intro.tex
\section{Introduction}\label{sec:intro}

Web agents—language models coupled with a browser and tool-use interfaces—have rapidly emerged as a core direction for building general-purpose assistants that can search, navigate, and act on the open web \citep{pmlr-v70-shi17a, liu2018reinforcement, zhouwebarena, huang-etal-2025-r2d2, 10.1145/3711896.3737384}. 
Despite striking demonstrations, progress is increasingly constrained by data. Existing benchmarks provide only human-annotated task descriptions paired with a curated end-to-end solution, offering no supervision over the agent’s latent decision process \citep{deng2023mind2web, Wang2023VoyagerAO, zhouwebarena, spangher-etal-2025-creative, huang2025teaching}. This is problematic: prior work has highlighted that web navigation is naturally a hidden Markov process, where the critical uncertainty lies in the unobserved intermediate states and actions \citep{sodhi2024step, huang-etal-2025-r2d2}. Without reliable ground-truth trajectories, agents are overfitting to small hand-authored sets, and struggling to generalize to realistic multi-hop, cross-page use cases \citep{Murty2025NNetNav, ishmam2026timewarp}.

To address the scarcity of large-scale data, recent efforts have begun to generate web tasks automatically. For instance, AgentTrek \citep{xu2024agenttrek} converts online tutorial documents into executable trajectories and validates them with a vision-language agent, while WebWalker \citep{wu2025webwalker} and NNetNav \citep{Murty2025NNetNav} use an LLM-based policy model to explore websites and then label the resulting trajectories with intent. While promising, these approaches face three critical limitations. First, they are \textbf{prohibitively expensive in both time and cost}: constructing a single task requires repeated agent--environment rollouts and multiple LLM calls, driving costs upward and limiting throughput. Second, they \textbf{induce strong exploration bias}: LLM policies tend to overfit to “obvious” paths (e.g., repeatedly selecting product detail pages), while overlooking more nuanced or less salient site functionalities such as rewards programs, sales, or new arrivals. As a result, large portions of the site graph remain unexplored and untested. Third, they \textbf{lack compositionality and utility}: most tasks generated by existing approaches collapse to single-hop retrieval or form-filling that could be solved by a simple search query, failing to capture the multi-hop, cross-page dependencies of real-world navigation. As a result, such tasks offer little practical utility for users, since solving them does not require deeper reasoning, sustained interaction, or problem-solving skills that would actually test or benefit an intelligent web agent in realistic settings. We provide detailed comparison in \Cref{tab:dataset_comparison}.


We introduce  \modelname (\texttt{\textbf{G}}enerate Long-Horizon \texttt{\textbf{T}}asks for Web \texttt{\textbf{A}}gents at
Scale), a scalable task-generation and benchmarking framework for web agents that integrates \emph{crawl}, \emph{retrieve}, \emph{in-context generation}, and \emph{quality control} to produce realistic, graded-difficulty tasks paired with \emph{executable ground-truth trajectories}. Concretely, we first crawl publicly accessible sites to construct a content/link graph; next, we retrieve candidate pages and evidence snippets as seeds; we then generate tasks via in-context prompting that references on-site facts, UI elements, and workflows; we filter with an LLM-based quality control protocol that enforces clarity and solvability against the current crawl; 
and finally we record a minimal, programmatically executable navigation path (clicks, scrolls, inputs) enabling deterministic replays and step-level attribution.

This design directly targets the three challenges identified earlier. \emph{Efficiency and cost}: we separate one-time crawling from lightweight, retrieval-seeded task generation, eliminating repeated policy rollouts and reducing both redundant LLM calls and wall-clock latency. \emph{Exploration coverage}: by leveraging retrieval-based seeding rather than policy-driven exploration, our method avoids the strong biases of LLM navigators and systematically samples diverse, semantically rich regions of each website’s graph. \emph{Compositionality and utility}: grounding generation in the site graph enables the construction of multi-hop tasks that require reasoning across interconnected pages, mirroring the dependencies and open-ended objectives characteristic of real user workflows.

\input{resources/benchmark_comparison}

To automatically generate diverse multi-hop web agent tasks, we distinguish between two primary settings: \textbf{intra-website queries}, where reasoning unfolds entirely within a single domain, and \textbf{inter-website queries}, which require coordination across multiple domains. Within intra-website tasks, our benchmark spans a broad range of languages, including English, Italian, Japanese, German, and Chinese, ensuring coverage beyond English-centric environments.

We instantiate the pipeline on over \textbf{50} public websites spanning e-commerce catalogs, entertainment/government pages, forums, and finances, etc. Beyond automatic filtering, we collect human annotations to assess answer correctness, complexity, utility and realism, confirming that tasks are reasonable for humans yet challenging for current agents. 
Our contributions are three-fold:
\begin{itemize}
     \item \textbf{Problem formulation.} We introduce and formally define the new task of \emph{multi-hop web-agent task generation},      
     where tasks require agents to reason across multiple pages or domains rather than collapsing to trivial information lookup. 

    \item \textbf{Efficient automatic data generation.} We propose an efficient pipeline that markedly reduces per-task time and cost while producing challenging tasks; each task includes an executable minimal ``gold path'' for deterministic replay and step-level diagnostics, safeguarded via verifier checks and negative controls.
  \item \textbf{Evolving benchmark.} Rather than treating \textsc{\modelname} as a fixed dataset, we frame it as a self-evolving benchmark ecosystem in which end users can continually generate tasks from current web content. Coupled with our verification pipeline,  \textsc{\modelname} preserves task quality and reproducibility while exposing agents to a changing, realistic task distribution, thereby reducing overfitting and enabling ongoing evaluation of generalization.

\end{itemize}

%% file: resources/benchmark_comparison.tex
\begin{table*}[t]
    \centering
    \small
    \renewcommand{\arraystretch}{1.0} 
    \begin{tabularx}{\textwidth}{l *{5}{>{\centering\arraybackslash}X}}
        \toprule
        Dataset & Multi-hop & Scalable Auto-gen & Groundtruth & Dynamic & Multilingual \\
        \midrule
        WebVoyager \cite{Wang2023VoyagerAO}  & \xmark & \xmark & \cmark & \xmark & \xmark \\
        WebArena  \cite{zhouwebarena}         & \xmark & \xmark & \cmark & \xmark & \xmark \\
        WebWalker \cite{wu2025webwalker}      & \cmark & \xmark & \cmark & \xmark & \cmark \\
        BrowseComp \cite{wei2025browsecomp} & \cmark & \xmark & \cmark & \xmark & \xmark \\
        WebDS \cite{hsu2025webdsendtoendbenchmarkwebbased} & \cmark & \xmark & \cmark & \xmark & \xmark \\
        AgentTrek \cite{xu2024agenttrek}      & \xmark & \cmark & \xmark & \xmark & \xmark \\
        NNetNav \cite{Murty2025NNetNav}       & \cmark & \cmark & \xmark & \xmark & \xmark \\
        \textbf{GTA} (Ours) & \cmark & \cmark & \cmark & \cmark & \cmark \\
        \bottomrule
    \end{tabularx}
    \caption{Comparison of benchmark properties. \textbf{GTA} is the only benchmark that combines 
    \emph{automatic task generation}, \emph{multi-hop reasoning}, \emph{executable ground-truth}, 
    \emph{dynamic expansion}, and \emph{multilingual coverage}.}
    \label{tab:dataset_comparison}
\end{table*}

%% file: sections/benchmark.tex
\section{Benchmark Design}
The goal of our benchmark is to create web-agent tasks that systematically capture the \emph{multi-hop} nature of realistic online interactions. Unlike prior datasets that treat each task as a linearized trajectory, we explicitly ground task generation in the structural properties of the underlying web graph. This design choice serves two objectives: (i) it reflects the fact that solving many real-world queries requires reasoning across multiple, interlinked webpages, and (ii) it enables reproducible task construction decoupled from transient agent rollouts. In what follows, we describe the full pipeline in four stages: collecting observations by crawling and constructing a site graph, generating multi-hop tasks through retrieval-augmented prompting, controlling task quality via systematic verification, and analyzing coverage relative to existing benchmarks.

\subsection{Benchmark Construction}
\stitle{Defining Multi-hop Tasks}
We aim to construct tasks that require \emph{multi-hop reasoning} over web environments.  
Formally, let $\mathcal{G}=(V,E)$ denote a directed web graph, where each node $v \in V$ corresponds to a crawled webpage and each edge $(v_i,v_j)\in E$ corresponds to a hyperlink from $v_i$ to $v_j$.  
A task is represented as $T = \langle q, A, P^\star \rangle$, where $q$ is the natural-language query, $A$ is the answer, and $P^\star = \langle v_1, v_2, \ldots, v_m \rangle$ is a minimal executable path within $\mathcal{G}$ that leads to $A$.

We define a task as \emph{multi-hop of order $n$} if solving it requires aggregating information from at least $n$ distinct webpages, i.e.,
\[
| \{ v \in P^\star : \text{evidence}(v) \neq \emptyset \} | \geq n, \quad n > 1.
\]
Here, $\text{evidence}(v)$ denotes the subset of page content from node $v$ that is necessary to derive the correct answer.  
In contrast, a single-hop task admits $n=1$, solvable by local reasoning from one node only.

\stitle{Collecting Observations}
To instantiate $\mathcal{G}$, we perform breadth-first crawling over publicly accessible domains. Each hyperlink discovered from page $v$ is initially added as a child node. However, paths to a given webpage are not unique, which risks cycles and inflated traversal lengths. To address this, we enforce a canonical shortest-path policy: whenever a node is reached via a shorter path, we update its parent assignment accordingly. Specifically, each node stores the accessibility tree of its corresponding webpage.

\stitle{Generating Multi-hop Tasks}
Once $\mathcal{G}$ is collected, we embed each node $v \in V$ into a dense representation $h_v \in \mathbb{R}^d$ using an LLM-based encoder $f_\theta$. Given a seed node $v_s$, we retrieve a candidate set $\mathcal{R}(v_s) = \{v_1,\ldots,v_k\}$ via dense retrieval. 
To induce compositionality, we randomly select a secondary node $v_t \in \mathcal{R}(v_s)$ and provide $(\text{content}(v_s), \text{content}(v_t))$ as context to a generator LLM.  
The model is prompted to synthesize a query $q$ whose answer $A$ requires reasoning across both pages. This retrieval-and-pairing strategy enforces multi-hop dependencies by design, rather than relying on incidental cross-references.

\stitle{Controlling Quality}
All generated tasks are subjected to a multi-stage verification pipeline:  

\begin{enumerate}[nosep,leftmargin=*]
    \item \emph{Multi-hop validation.} We reject cases where the model produces a conjunction of two independent single-hop queries (e.g., linked by ``and''), which do not constitute genuine compositional reasoning.
    \item \emph{Answer correctness.} For each candidate $(q, A)$, we condition a verifier LLM on the content of supporting nodes only, requiring it to reproduce or validate $A$. Incorrect or unverifiable outputs are discarded.
    \item \emph{Ambiguity detection.} We assess whether $q$ admits a unique answer. Queries missing critical qualifiers (e.g., temporal or geographic constraints) are filtered out.
    \item \emph{Solvability check.} We ensure the existence of at least one minimal executable path $P^\star$ within $\mathcal{G}$ such that traversal yields sufficient evidence to derive $A$. This guarantees deterministic replayability and prevents spurious generations.
\end{enumerate}

The resulting benchmark thus consists of well-formed, unambiguous multi-hop tasks, each paired with a cached graph $\mathcal{G}$ and a minimal executable trajectory $P^\star$, enabling reproducible training and evaluation.
\subsection{Benchmark Details}
A key feature of our pipeline is that it is \emph{fully automatic}: once a website graph is constructed, tasks can be generated without any manual authoring or intervention. This design implies that, in principle, the pipeline can scale to produce arbitrarily many tasks across arbitrarily many domains.

\stitle{Domains} In this paper, we instantiate the benchmark on over 50 publicly accessible websites spanning healthcare, finance, e-commerce, entertainment, government, and scientific resources. This breadth ensures heterogeneous evaluation settings, as summarized in \Cref{tab:domains}.

\stitle{Number of Tasks} In total, we generate \textit{5,000} intra-website tasks across the covered domains. Within the Healthcare and Finance group of websites, we generate \textit{600} cross-website inter-website tasks. These generated tasks are used for experiments in this paper.

\stitle{Pipeline} We release the generation pipeline to allow end users to create up-to-date tasks directly from live websites. This supports continual refresh of evaluation data, reduces overfitting to a static benchmark, and reframes \textsc{\modelname} as a reusable task-generation framework rather than a one-time dataset.

%% file: sections/analysis.tex
\section{Analysis}

In this section, we first compare query structures across datasets (\Cref{ssec: qual}). We then quantitatively assess task difficulty by benchmarking performance against a simple search-based baseline, revealing that, unlike previous datasets, \modelname tasks cannot be trivially solved by single-hop retrieval (\Cref{ssec: search_baseline}). We perform a page-coverage analysis to measure how broadly each benchmark explores the underlying website graph, showing that \textsc{\modelname} achieves substantially higher structural diversity (\Cref{ssec: page_coverage}). We present human evaluation results confirming that \textsc{\modelname} yields complex, verifiable, and realistic tasks for web-agent assessment (\Cref{ssec: human_eval}). Finally, we present a time and cost efficiency analysis (\Cref{ssec: efficiency}).

\input{resources/example_tasks_table}

\subsection{Qualitative Comparison of Queries}
\label{ssec: qual}
A closer examination of representative tasks highlights substantial differences between \textsc{\modelname} and prior benchmarks. Unlike WebDS, which largely consists of factual lookups or pairwise comparisons (e.g., retrieving COVID-19 statistics or population growth rates), \textsc{\modelname} tasks are explicitly compositional. They often require integrating multiple heterogeneous sources of evidence, such as combining numerical statistics (career strikeouts), product attributes (price and ratings), and domain-specific safety constraints (pharmacological dosage and drug interactions). This design ensures that tasks involve multi-hop reasoning rather than collapsing into atomic queries.  

Another distinction lies in the nature of the required reasoning. \textsc{\modelname} emphasizes \emph{context-dependent decision-making}, where answers are not limited to factual retrieval but also demand comparative or evaluative judgments (e.g., identifying which product is superior, or whether a medical combination is safe). By contrast, AgentTrek and NNetNav focus on navigation-oriented instructions whose outcomes are often underspecified (e.g., ``expand results on SeatGeek'' or ``find a restaurant'').

Finally, the answer space of \textsc{\modelname} tasks is designed to be precise and verifiable, grounded in structured numerical, categorical, or medical evidence. This stands in contrast to the open-ended or underspecified nature of prior benchmarks, where the lack of explicit answers limits their utility for rigorous evaluation. Collectively, these properties position \textsc{\modelname} as a benchmark that emphasizes \emph{long-horizon, evidence-grounded task construction}, thereby enabling a more faithful assessment of reasoning and integration capabilities in web agents.

\input{resources/google_hitrate_fig}

\subsection{Difficulty Comparison Against a Simple Search Baseline}
\label{ssec: search_baseline}
We first evaluate a simple baseline where an LLM retrieves results from the Google Search API and generates an answer based on the returned snippets. To enable a fair comparison, we filter information-retrieval style tasks from both AgentTrek and NNetNav. As shown in \Cref{fig:page_coverage_rate}, the correctness rate is surprisingly high: 95\% on AgentTrek and 100\% on NNetNav. On \modelname, however, the baseline only reaches 14\%. This indicates that tasks in AgentTrek and NNetNav often reduce to single-hop fact lookup, which can be trivially solved by a search engine. The inflated correctness rates reveal that these datasets do not capture the core challenge of web agents: reasoning over dispersed information across multiple pages and interfaces. 

\subsection{Page-Coverage Analysis}
\label{ssec: page_coverage}
We next turn to the \emph{page-coverage rate}, defined as the fraction of first-level nodes in the crawled website graph (roughly corresponding to distinct site functionalities, such as “team roster,” “product details,” or “checkout page”) that are exercised by benchmark tasks. This metric is inspired by the \emph{line-coverage} concept in programming languages, which measures how much of the codebase is actually exercised by a test suite. Analogously, high page coverage implies that benchmark tasks probe a diverse set of site functionalities, while low coverage indicates narrow task construction.

\input{resources/page_coverage}

As shown in \Cref{fig:page_coverage_rate}, the coverage is extremely limited: 20.2\% for AgentTrek, 24.3\% for NNetNav, and only 11.0\% for WebDS, which are all far lower than \modelname page coverage rate.
This narrow span suggests that existing datasets focus disproportionately on a small subset of site functionality. 
The overlooked regions represent critical parts of the user experience, yet they remain underrepresented in benchmark tasks.
Taken together, these analyses demonstrate that prior benchmarks (i) collapse to factoid-style queries easily solved by search engines, and (ii) induce strong exploration biases that leave large swaths of website functionality untested. 

\subsection{Human Evaluation}
\label{ssec: human_eval}
\stitle{Correctness} To further assess the reliability of our automatically constructed benchmark, we randomly sample 100 tasks and evaluate the correctness of the annotated answers through human judgment. For each sampled task, we provide annotators with the generated task description, the proposed answer, and the two source web pages from which the answer is derived. Annotators are asked to verify whether the annotated answer is indeed correct given the provided evidence. The resulting human agreement rate reaches 87\%, demonstrating that our pipeline produces tasks and answers that are both interpretable and verifiable, with only a small fraction of cases requiring clarification.

\stitle{Complexity, Utility, Realism and Quality}  
We further conduct a controlled human study (20 tasks per dataset) to compare tasks generated by our approach (\textsc{\modelname}), WebDS, and AgentTrek along four dimensions: complexity, utility, realism, and overall quality. Results are shown in Table~\ref{tab:human-eval}. 

\input{resources/human_eval}

Annotators judge \textsc{\modelname} tasks to be substantially more complex (\textbf{2.35}) than those in WebDS (1.90) and AgentTrek (1.23), indicating that our pipeline successfully produces multi-hop navigation and reasoning scenarios rather than trivial, single-hop lookups. On utility, \textsc{\modelname} achieves the highest rating (\textbf{2.03}), followed closely by AgentTrek (2.00) and WebDS (1.98), suggesting that \textsc{\modelname} maintains comparable usefulness with added reasoning depth.  

For realism, AgentTrek scores the highest (\textbf{1.98}), outperforming WebDS (1.68) and \textsc{\modelname} (1.53). This gap may stem from our graph-grounded generation process, which can introduce queries that are less reflective of natural user intents. Finally, \textsc{\modelname} attains the best overall quality score (\textbf{1.82}), surpassing both WebDS (1.77) and AgentTrek (1.73), highlighting that tasks generated by our approach are both coherent and challenging.  

Taken together, these results demonstrate that \modelname yields tasks that best balance realism, reasoning depth, and evaluation value—producing scenarios that are both practically useful and more faithful to the complexities of real-world web navigation.

\subsection{Time and Cost Efficiency}
\label{ssec: efficiency}
A key advantage of \textsc{\modelname} lies in its scalability and efficiency. We present detailed quantitative comparisons in \Cref{app:cost}. In summary, \textsc{\modelname} achieves a dramatic reduction in both computational and financial overhead compared to prior task-generation pipelines.

%% file: resources/example_tasks_table.tex
\begin{table*}[ht!]
\centering
\scriptsize
\renewcommand{\arraystretch}{1.2}
\begin{tabularx}{\textwidth}{c|X|X}
\toprule
\textbf{Dataset} & \textbf{Task} & \textbf{Answer / Outcome} \\
\midrule
\multirow{9}{*}{GTA} 
 & Who has more career strikeouts in their MLB career, Derek Hill or Caleb Ferguson? & Caleb Ferguson has more career strikeouts (362). \\
\cmidrule{2-3}
 & What is the total combined price of the UA Blur Pro Men's Football Cleats and the UA Spotlight Pro Men's Football Cleats, and which of these two products has the higher overall customer rating? & The total combined price is \$240 (\$110 for UA Blur Pro and \$130 for UA Spotlight Pro). The UA Spotlight Pro Men's Football Cleats has the higher overall customer rating of 4.7 compared to 4.5 for the UA Blur Pro Men's Football Cleats. \\
\cmidrule{2-3}
 & What is the maximum single dose of diphenhydramine (in mg) recommended for insomnia in adults, and is it safe to take this dose concurrently with sodium oxybate according to the interaction warnings? & The maximum single dose of diphenhydramine recommended for insomnia in adults is 76 mg (as diphenhydramine citrate), and it is not safe to take this dose concurrently with sodium oxybate due to potential additive CNS depression effects. \\
\midrule
\multirow{3}{*}{WebDS} 
 & Which currency pair (GBP/USD, EUR/JPY, USD/AUD) had the largest fluctuation in past 72h? & EUR/JPY had the largest fluctuation (high: 163.81 JPY). \\
\cmidrule{2-3}
 & Compare 2020 population growth rates of Brazil vs.\ Indonesia. & Brazil: 0.72\%. Indonesia: 1.07\%. Indonesia higher. \\
\cmidrule{2-3}
 & Total confirmed COVID-19 cases in California (2022). & 5,634,397 cases. \\
\midrule
\multirow{3}{*}{AgentTrek} 
 & Explore Wikipedia's ``Did you know?'' section. & N/A \\
\cmidrule{2-3}
 & Search for family events in Miami, FL and expand results on SeatGeek. & N/A \\
\midrule
\multirow{3}{*}{NNetNav} 
 & Find healthy dinner ideas. & N/A \\
\cmidrule{2-3}
 & Find event spaces in San Francisco, CA. & N/A \\
\bottomrule
\end{tabularx}
\caption{Example web tasks sampled from four datasets: GTA, WebDS, AgentTrek, and NNetNav. Each row shows a representative task and its summarized answer or outcome.}
\label{tab:webtasks}
\end{table*}

%% file: resources/google_hitrate_fig.tex


%% file: resources/page_coverage.tex
\begin{figure}[h]
    \centering
    \footnotesize
    \includegraphics[width=0.95\linewidth]{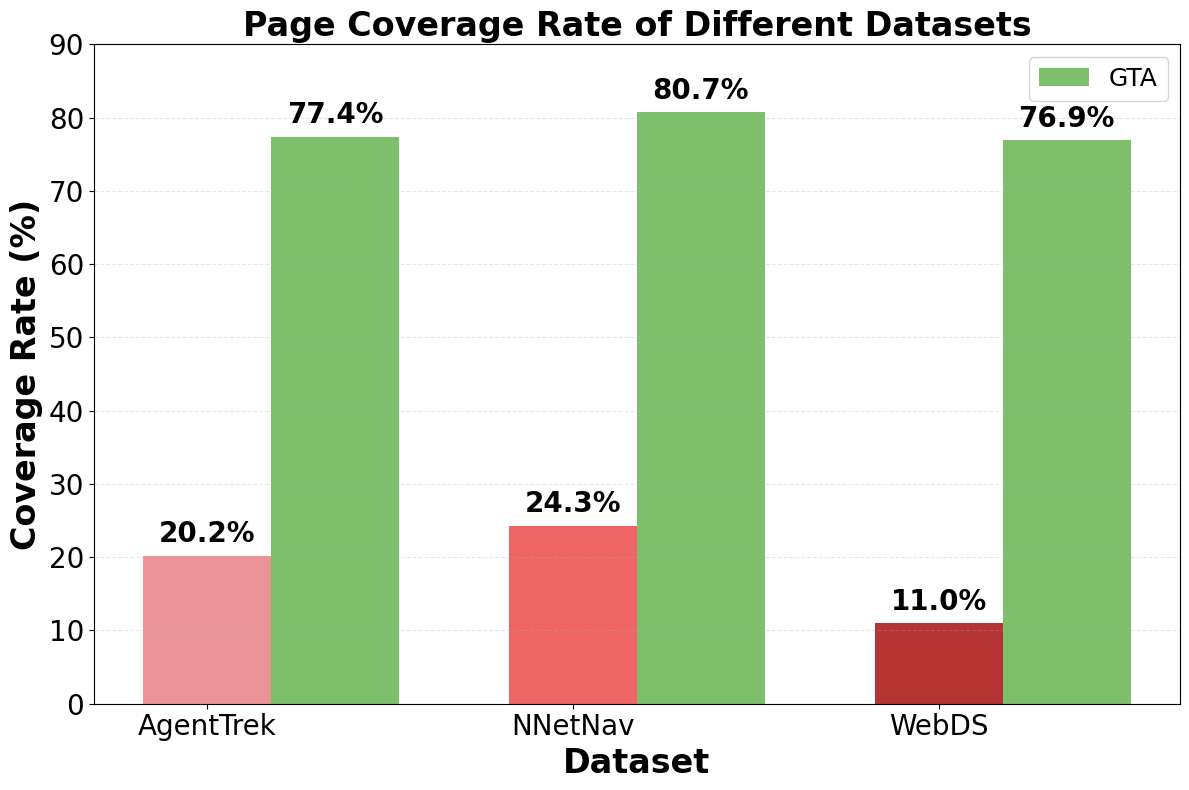}
    \caption{Page coverage rate comparison across datasets. Our proposed benchmark (\textsc{GTA}) achieves substantially higher page coverage (77.4\%--80.7\%) compared to prior datasets such as AgentTrek (20.2\%), NNetNav (24.3\%), and WebDS (11.0\%). This indicates that \textsc{GTA} tasks exercise a much broader range of site functionalities, reflecting stronger information-sourcing ability. Statistics calculated on \texttt{espn.com} for AgentTrek, \texttt{underarmour.com} for NNetNav and \texttt{casper.com} for WebDS.}
    \label{fig:page_coverage_rate}
    \vspace{-3mm}
\end{figure}

%% file: resources/human_eval.tex
\begin{table}[h!]
\centering
\footnotesize
\setlength{\tabcolsep}{4pt}
\renewcommand{\arraystretch}{1.0}
\begin{tabularx}{\columnwidth}{lcccc}
\toprule
\textbf{Dataset} & \textbf{Complexity} & \textbf{Utility} & \textbf{Realism} & \textbf{Quality} \\
\midrule
\textsc{WebDS} & 1.90 & 1.98 & 1.68 & 1.77 \\
\textsc{AgentTrek} & 1.23 & 2.00 & \textbf{1.98} & 1.73 \\
\textsc{GTA} & \textbf{2.35} & \textbf{2.03} & 1.53 & \textbf{1.82} \\
\bottomrule
\end{tabularx}
\vspace{-2mm}
\caption{Human evaluation (Likert 1–3) of task complexity, utility, realism, and overall quality. The highest value per column is bolded.}
\label{tab:human-eval}
\vspace{-3mm}
\end{table}

%% file: sections/results.tex
\section{Experiments}
In this section, we first introduce the baseline agents evaluated in this paper (\Cref{ssec: baseline}).
We then report their performance on \modelname intra- and inter-website tasks (\Cref{ssec: benchmark_results}).
We also extend evaluation to multilingual web environments (\Cref{ssec: multilingual_results}).
Finally, we conduct an error analysis to identify key failure modes (\Cref{ssec: error_analysis}). Specific implementation details are presented in \Cref{app: implementation}.

\subsection{Baselines}
\label{ssec: baseline}
\stitle{Browser Use}\footnote{\url{https://github.com/browser-use/browser-use}.} Browser Use is an open-source framework that couples large language models with a real browser environment. The perceive–decide–act cycle enables robust handling of multi-step workflows such as navigation, form-filling, and information extraction across diverse sites. It is a widely adopted baseline, achieving strong results on benchmarks like WebVoyager.

\stitle{AgentOccam} \cite{yangagentoccam} AgentOccam represents a complementary approach that prioritizes simplicity and alignment. It refines the interface between the LLM and the web environment by pruning rarely used actions and filtering page observations into concise textual representations. It has achieved state-of-the
art performance on benchmarks like WebArena.

We evaluate a limited number of agent backbones by design, consistent with prior web-agent benchmarks, which typically focus on representative agent paradigms rather than exhaustive architectural sweeps, for example, WebDS\cite{hsu2025webdsendtoendbenchmarkwebbased} evaluates Browser  Use and AgentOccam, while NNetNav\cite{Murty2025NNetNav} and WebWalker\cite{wu2025webwalker} primarily study ReAct-style agents. Our goal is to validate the benchmark under commonly used agent behaviors; accordingly, the primary contribution lies in the benchmark itself rather than in the breadth of agent architectures evaluated.


\subsection{Benchmark Results}
\label{ssec: benchmark_results}
\stitle{Intra-website Task Performance} \Cref{fig:main_result} reports success rates of Browser Use and AgentOccam across domains.
A clear gap emerges between our benchmark and prior ones such as WebVoyager and Mind2Web. While both agents achieve relatively high scores on these earlier datasets (up to 82\% on WebVoyager and 45\% on Mind2Web), their performance on our benchmark domains is markedly lower, often below 20--30\%. This contrast suggests that existing benchmarks have become static and saturated, allowing agents to overfit to templated interfaces or repeated patterns, whereas our benchmark introduces dynamic, evolving sites that demand genuine adaptation and remain useful over a longer horizon.

Beyond the overall drop in success rate, we also observe sharp performance variations across domains. As shown in \Cref{fig:performance_per_website} 
\texttt{musicbrainz.org} or \texttt{cvshealth.com} admit moderate success, whereas shopping portals like \texttt{underarmour.com} and \texttt{ticketcenter.com} prove substantially harder. Such heterogeneity highlights that real-world navigation cannot be captured by single-domain evaluation: robustness requires agents to generalize across diverse layouts, workflows, and interaction styles.

\input{resources/main_result}

\stitle{Inter-website Task Performance} To demonstrate the flexibility of our pipeline, we also generate \emph{inter-website multi-hop tasks}, where solving a task requires integrating information from multiple websites rather than staying within a single domain. For these experiments, we select two domains with naturally complementary sources of information: healthcare (Mayo Clinic, WebMD, Drugs.com, CVS Health, and CDC) and finance (Yahoo Finance, markets.ft and Seeking Alpha).

We evaluate these tasks using the Browser Use agent, which executes tasks in a real browser environment. Experiments show that success rates for cross-website tasks are substantially lower than those observed in single-website settings. For example, Mayo Clinic and WebMD tasks achieve a 12.5\% success rate, CDC and WebMD reach 8.2\%, and Drugs.com \& CVS Health drop to just 4.2\%. These numbers are considerably lower than the success rate of same-website tasks reported earlier, confirming that cross-website integration poses unique challenges.

We attribute this difficulty to two factors. First, cross-website tasks require agents to reason about \emph{complementary but non-overlapping knowledge}: one site may provide background definitions or guidelines, while the other presents instance-specific details (e.g., drug side effects vs. dosage instructions). Second, agents must manage \emph{distributional shifts across sites}, such as inconsistent terminology, different UI structures, or divergent information formats. This stands in contrast to same-site navigation, where once an agent adapts to the local interface, it can generalize within that site.


\input{resources/cross_web_results}

\subsection{Multi-lingual Task Performance}
\label{ssec: multilingual_results}
To further evaluate the flexibility and coverage of our pipeline, we extend task generation beyond English to \emph{multilingual web environments} spanning the \textit{health}, \textit{sports}, and \textit{news} domains. We crawled representative websites in four languages—English, Italian, German, and Japanese—and constructed tasks from each. We then evaluate these tasks using two representative web agents, Browser Use and AgentOccam. As shown in \Cref{fig:multilingual}, the results reveal a pronounced degradation in performance on non-English websites. For example, the Browser Use agent attains a 0\% success rate on the German site, 2.9\% on the Japanese site, and 0\% on the Italian site. In contrast, the same agent achieves markedly higher performance on English-language health websites, with success rates of 8.2\% on CDC and 6.0\% on NIH.gov.
These findings highlight a strong bias in current web agents toward English-centric environments, while exposing severe limitations in their ability to operate on foreign-language websites. A similar trend is consistently observed with AgentOccam and across other domains, corroborating the language-specific performance disparity.


\subsection{Error Analysis}

We conduct a detailed error analysis on a random sample of 100 failed tasks to better understand the limitations of current web agents on our benchmark. Three major categories of failure modes emerge.

\stitle{Failure to Reach All Required Webpages (90\%)}  
\label{ssec: error_analysis}
A common error pattern arises when the agent fails to correctly navigate to all webpages required by a multi-hop query. Since many tasks require synthesizing evidence from two or more distinct sites, failure to reach even one of them leads to incomplete reasoning chains and ultimately incorrect answers. We observe that this issue often stems from brittle link-following strategies, where the agent either misinterprets hyperlink semantics (e.g., clicking on visually similar but contextually irrelevant links) or becomes trapped in navigation loops. 

\stitle{Early Stopping (40\%)}  
Another prominent failure mode involves premature termination of the task. Agents sometimes generate a final answer after retrieving partial evidence, without fully verifying consistency across all required sources. This behavior is especially prevalent in tasks that involve numerical aggregation or comparisons, where one piece of evidence might seem sufficient but is actually incomplete. Such early stopping reflects the broader challenge of balancing efficiency with completeness in long-horizon reasoning. 

\stitle{Over-reliance on Search Box Operations (40\%)}  
We also find that agents frequently overuse on-site search boxes rather than exploiting the explicit structure of the web graph. While search boxes sometimes provide shortcuts, they are not guaranteed to expose the necessary evidence, and reliance on them can bypass important compositional navigation steps. In many cases, agents retrieve irrelevant or overly broad results through poorly formulated queries, which then propagate errors into the reasoning process. 


%% file: resources/main_result.tex
\begin{figure*}[t!]
    \centering
    \footnotesize
    \includegraphics[width=0.8\linewidth]{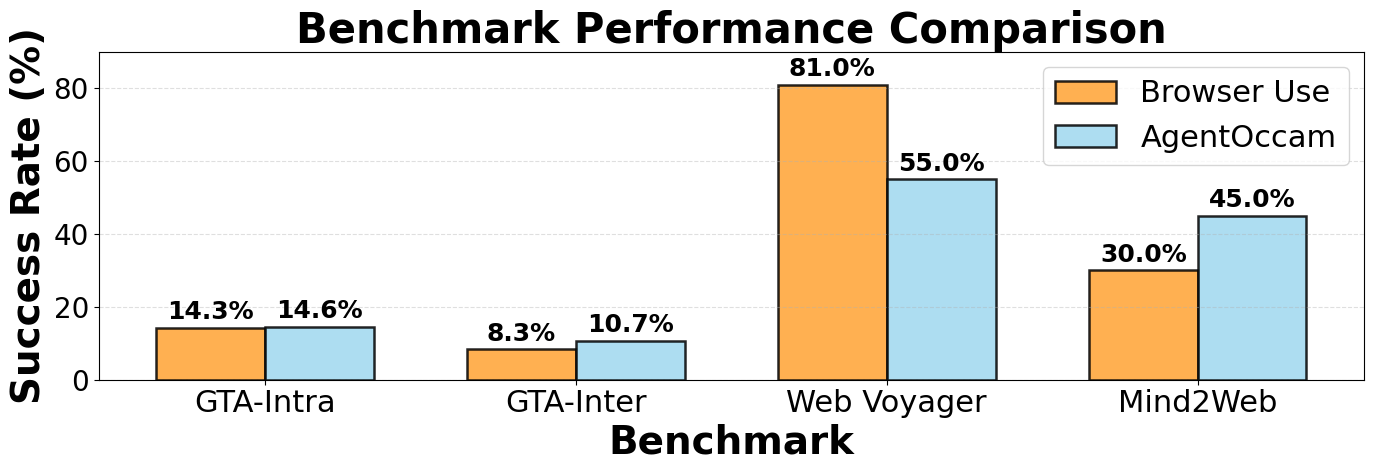}
    \caption{Benchmark performance comparison of Browser Use and AgentOccam. 
On GTA (intra/inter-website) tasks, both agents achieve relatively low success rates, highlighting the difficulty of 
multi-hop, compositional reasoning. By contrast, performance is saturated on 
WebVoyager and Mind2Web. Human performance on GTA tasks is 85\%.}
    
    \label{fig:main_result}
    \vspace{-5mm}
\end{figure*}

%% file: resources/cross_web_results.tex

%% file: sections/related_works.tex
\section{Related Work}

A number of benchmarks have been developed to evaluate web agents. Mind2Web \cite{deng2023mind2web} provides human-authored task--solution pairs for open-domain web navigation. WebVoyager \cite{Wang2023VoyagerAO}, WebArena \cite{zhouwebarena}, BrowseComp \cite{wei2025browsecomp} and WebDS \cite{hsu2025webdsendtoendbenchmarkwebbased} provide realistic environments and curated tasks for web-based decision making. However, these benchmarks mainly provide end-to-end trajectories, offering limited support for fine-grained diagnostics and compositional reasoning.

To reduce annotation costs, recent work has explored automatic task generation. AgentTrek \cite{xu2024agenttrek} transforms web tutorials into executable trajectories, while WebWalker \cite{wu2025webwalker} and NNetNav \cite{Murty2025NNetNav} use LLM-driven exploration to synthesize tasks from raw environments. Despite their promise, these approaches can be expensive due to repeated rollouts, biased toward salient regions of a site, and often produce tasks that collapse into simple single-hop retrieval.
\modelname also differs in construction strategy. Unlike WebWalker, which discovers tasks through model-driven exploration, \modelname first systematically crawls websites, builds explicit site graphs, and then generates tasks from enumerated graph structures. This design reduces exploration bias and enables controlled graph coverage and cross-page evidence composition. Compared with human-authored benchmarks such as BrowseComp, \modelname is fully automated and scalable across 50+ websites, while explicitly guaranteeing multi-hop evidence distribution, cross-page dependency, and non-answerability from any single page.

%% file: sections/conclusion.tex
\section{Conclusion}

We introduced \textsc{\modelname}, a scalable benchmark for web agents. 
By decoupling crawling from task generation and anchoring tasks to the site graph, \textsc{\modelname} produces compositional, verifiable tasks with executable trajectories across diverse domains and languages. 

Our analyses show that \textsc{\modelname} surfaces challenges overlooked by prior benchmarks—agents fail on cross-website and multilingual tasks, and simple search baselines cannot solve them. Human studies further confirm that tasks are more complex, useful, and diagnostic. We hope \textsc{\modelname} will serve as a foundation for advancing robust, generalizable web agents.

\section*{Limitations}

\stitle{Scope of Website Coverage}
Although GTA spans over 50 publicly accessible websites across diverse domains and languages, the current selection remains limited relative to the vast heterogeneity of the modern web. Certain interaction types—such as authentication-gated services, dynamically rendered JavaScript interfaces, and transaction-based workflows—are excluded due to ethical and security constraints. Extending GTA to such settings will require robust sandboxing and privacy-preserving mechanisms for exploration.

\stitle{Quality Verification and Reasoning Depth}
The automatic verification process, though effective at filtering low-quality or ambiguous tasks, still depends on LLM-based validators. Subtle reasoning errors or underspecified compositional dependencies may occasionally persist. Incorporating human-in-the-loop validation or multi-model consensus could further strengthen quality assurance, especially for tasks involving nuanced domain knowledge.

\stitle{Dependence on Crawl Snapshots}
Our pipeline relies on periodic crawls to construct site graphs, enabling scalable data collection and timely iteration. However, task difficulty may vary due to multiple factors, including structural updates in websites \cite{ishmam2026timewarp}, the time interval between task creation and evaluation, and content drift or removal. These dynamics can affect task solvability and consistency across different crawl snapshots. Future work should explore more generalizable web agents that are robust to such temporal and structural variations. 

\stitle{Scope of Task Types}
\modelname focuses on multi-hop information-seeking tasks rather than the full range of web agent behaviors. In particular, it does not cover procedural workflows such as form-filling, checkout, or other action-heavy interactions. This scope is intentional: prior information-seeking benchmarks often collapse into single-hop retrieval, leaving cross-page evidence integration underexplored. Procedural workflows are an important complementary challenge and are already represented in benchmarks such as WebArena.


\section*{Ethics Statement}
We strictly adhere to ethical and legal standards in data collection and experimental design. All web data used in this study were obtained through automated scripts that fully comply with each website’s crawler and robots.txt policies. We only accessed publicly available webpages and refrained from interacting with or collecting any user-specific or private information. The crawling process respected server load limitations by introducing randomized delays and rate-limiting to minimize network impact.  

Additionally, all data were used solely for research purposes and are released or analyzed in aggregate form to prevent any potential misuse or privacy risk. Our benchmark construction and experiments comply with institutional and community ethical guidelines for web data research.

\section*{Acknowledgments}

Muhao Chen was supported in part by the National Science Foundation under grants OAC-2531126 and ITE-2333736.

%% file: resources/webiste_coverage_table.tex
\begin{table*}[h]
\centering
\scriptsize
\renewcommand{\arraystretch}{1.2}

\newcolumntype{Y}{>{\centering\arraybackslash}X}

\begin{tabularx}{\textwidth}{l|Y}
\toprule
\textbf{Domain} & \textbf{Websites} \\
\midrule
Healthcare and Medicine & 
\texttt{nih.gov}, \texttt{cdc.gov}, \texttt{mayoclinic.org}, \texttt{webmd.com}, 
\texttt{drugs.com}, \texttt{pubmed.ncbi.nlm.nih.gov}, \texttt{cvshealth.com}, 
\texttt{mhlw.go.jp}, \texttt{salute.gov.it}, \texttt{nhc.gov.cn}, 
\texttt{health.people.com.cn}, \texttt{bundesgesundheitsministerium.de}, 
\texttt{sante.gouv.fr} \\
\midrule
Finance and Economics & 
\texttt{bloomberg.com}, \texttt{finance.yahoo.com}, \texttt{markets.ft.com}, 
\texttt{seekingalpha.com}, \texttt{morningstar.com}, \texttt{bea.gov}, 
\texttt{stlouisfed.org}, \texttt{fred.stlouisfed.org}, \texttt{tradingeconomics.com}, 
\texttt{consumerfinance.gov}, \texttt{sec.gov}  \\
\midrule
E-commerce and Consumer Platforms & 
\texttt{underarmour.com}, \texttt{casper.com}, \texttt{resy.com}, 
\texttt{carnival.com}, \texttt{expedia.com}, \texttt{yellowpages.com} \\
\midrule
News, Entertainment, and Social Media & 
\texttt{espn.com}, \texttt{pro-football-reference.com}, \texttt{fbref.com}, 
\texttt{last.fm}, \texttt{musicbrainz.org}, \texttt{ticketcenter.com}, 
\texttt{tripadvisor.com}, \texttt{facebook.com} \\
\midrule
Government, Science, and Public Resources & 
\texttt{noaa.gov}, \texttt{climate.gov}, \texttt{nps.gov}, \texttt{mbta.com}, 
\texttt{new.mta.info}, \texttt{dmv.virginia.gov}, \texttt{iata.org}, 
\texttt{worldpop.org}, \texttt{ourworldindata.org}, \texttt{statista.com}, 
\texttt{riaa.com}, \texttt{ir.mit.edu} \\
\bottomrule
\end{tabularx}
\caption{Domain coverage of our benchmark. Sites span healthcare, finance, e-commerce, entertainment, and government/science, ensuring diverse and multilingual evaluation settings.}
\label{tab:domains}
\end{table*}

%% file: resources/prompts/task_generation_prompt.tex
\begin{figure}[h]
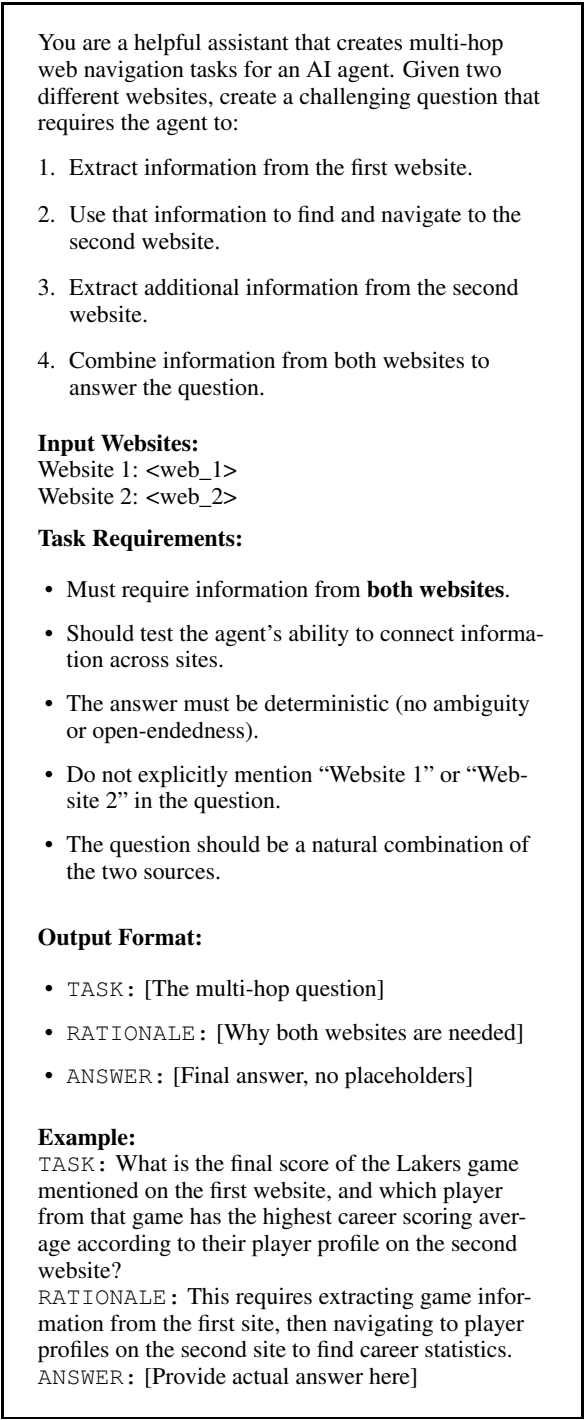

\centering
\begin{tcolorbox}[promptbox,title=]
\small
\RaggedRight
You are a helpful assistant that creates multi-hop web navigation tasks for an AI agent.  
Given two different websites, create a challenging question that requires the agent to:

\begin{enumerate}[leftmargin=*,itemsep=3pt,topsep=6pt]
  \item Extract information from the first website.
  \item Use that information to find and navigate to the second website.
  \item Extract additional information from the second website.
  \item Combine information from both websites to answer the question.
\end{enumerate}

\vspace{0.4\baselineskip}
\textbf{Input Websites:}\\
Website 1: \textless web\_1\textgreater \\
Website 2: \textless web\_2\textgreater

\vspace{0.6\baselineskip}
\textbf{Task Requirements:}
\begin{itemize}[leftmargin=1.2em,itemsep=2pt]
  \item Must require information from \textbf{both websites}.  
  \item Should test the agent’s ability to connect information across sites.  
  \item The answer must be deterministic (no ambiguity or open-endedness).  
  \item Do not explicitly mention “Website 1” or “Website 2” in the question.  
  \item The question should be a natural combination of the two sources.  
\end{itemize}

\vspace{0.6\baselineskip}
\textbf{Output Format:}
\begin{itemize}[leftmargin=1.2em,itemsep=2pt]
  \item \texttt{TASK:} [The multi-hop question]  
  \item \texttt{RATIONALE:} [Why both websites are needed]  
  \item \texttt{ANSWER:} [Final answer, no placeholders]  
\end{itemize}

\vspace{0.4\baselineskip}
\textbf{Example:}\\
\texttt{TASK:} What is the final score of the Lakers game mentioned on the first website, and which player from that game has the highest career scoring average according to their player profile on the second website?\\
\texttt{RATIONALE:} This requires extracting game information from the first site, then navigating to player profiles on the second site to find career statistics.\\
\texttt{ANSWER:} [Provide actual answer here]
\end{tcolorbox}
\captionsetup{font=small}
\caption{Prompt template for generating multi-hop web navigation tasks.}
\label{fig:multihop_prompt}
\end{figure}

%% file: resources/prompts/quality_control_prompts.tex
\begin{figure}[h]
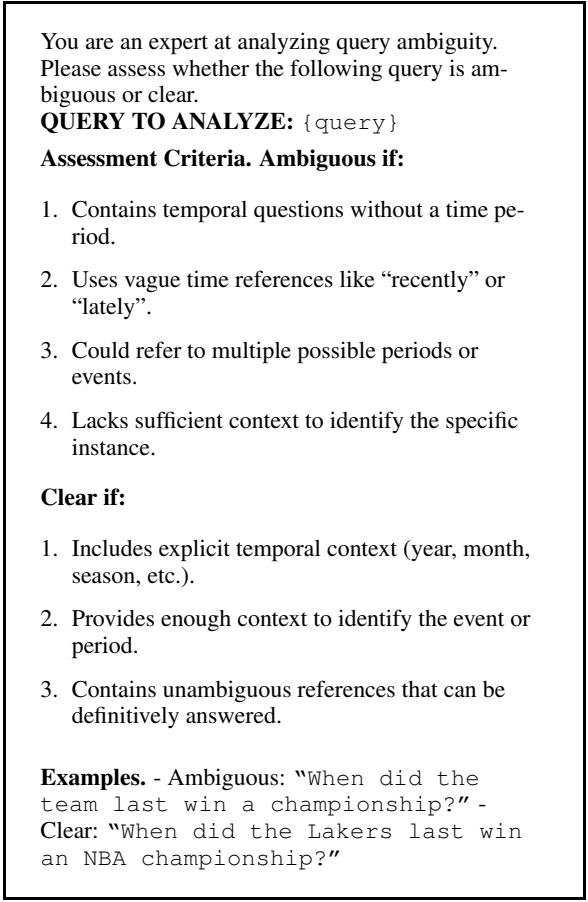

\centering
\begin{tcolorbox}[promptbox,title=]
\small
\RaggedRight
You are an expert at analyzing query ambiguity. Please assess whether the following query is ambiguous or clear.  

\textbf{QUERY TO ANALYZE:} \texttt{\{query\}}  

\stitle{Assessment Criteria}  
\textbf{Ambiguous if:}
\begin{enumerate}[leftmargin=*,itemsep=2pt]
  \item Contains temporal questions without a time period.  
  \item Uses vague time references like ``recently'' or ``lately''.  
  \item Could refer to multiple possible periods or events.  
  \item Lacks sufficient context to identify the specific instance.  
\end{enumerate}

\textbf{Clear if:}
\begin{enumerate}[leftmargin=*,itemsep=2pt]
  \item Includes explicit temporal context (year, month, season, etc.).  
  \item Provides enough context to identify the event or period.  
  \item Contains unambiguous references that can be definitively answered.  
\end{enumerate}

\stitle{Examples}  
- Ambiguous: \texttt{``When did the team last win a championship?''}  
- Clear: \texttt{``When did the Lakers last win an NBA championship?''}  

\end{tcolorbox}
\caption{Prompt for ambiguity assessment of queries.}
\label{fig:ambiguity_assessment}
\end{figure}

\begin{figure}[h]
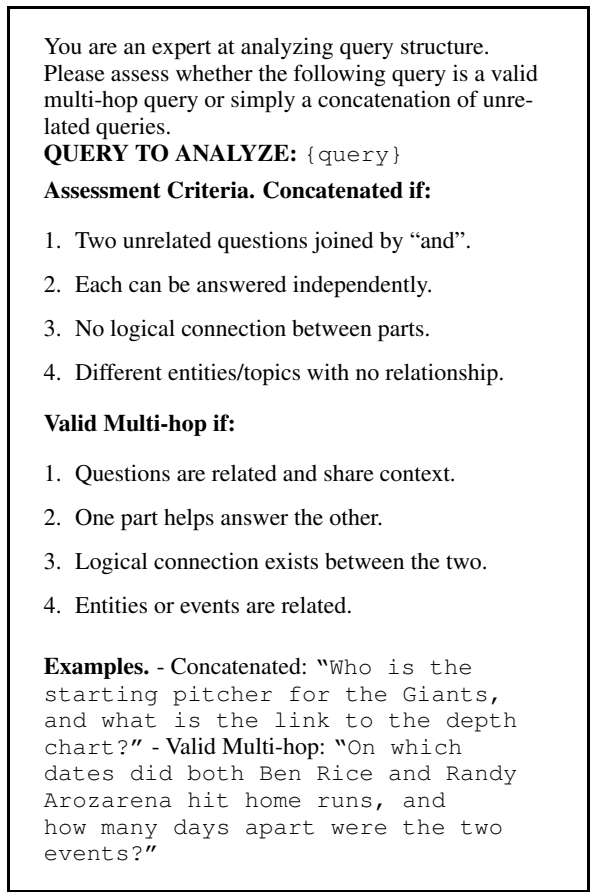

\centering
\begin{tcolorbox}[promptbox,title=]
\small
\RaggedRight
You are an expert at analyzing query structure. Please assess whether the following query is a valid multi-hop query or simply a concatenation of unrelated queries.  

\textbf{QUERY TO ANALYZE:} \texttt{\{query\}}  

\stitle{Assessment Criteria}  
\textbf{Concatenated if:}
\begin{enumerate}[leftmargin=*,itemsep=2pt]
  \item Two unrelated questions joined by ``and''.  
  \item Each can be answered independently.  
  \item No logical connection between parts.  
  \item Different entities/topics with no relationship.  
\end{enumerate}

\textbf{Valid Multi-hop if:}
\begin{enumerate}[leftmargin=*,itemsep=2pt]
  \item Questions are related and share context.  
  \item One part helps answer the other.  
  \item Logical connection exists between the two.  
  \item Entities or events are related.  
\end{enumerate}

\stitle{Examples}  
- Concatenated: \texttt{``Who is the starting pitcher for the Giants, and what is the link to the depth chart?''}  
- Valid Multi-hop: \texttt{``On which dates did both Ben Rice and Randy Arozarena hit home runs, and how many days apart were the two events?''}

\end{tcolorbox}
\caption{Prompt for distinguishing concatenated vs valid multi-hop queries.}
\label{fig:concatenation_assessment}
\end{figure}

\begin{figure}[h]
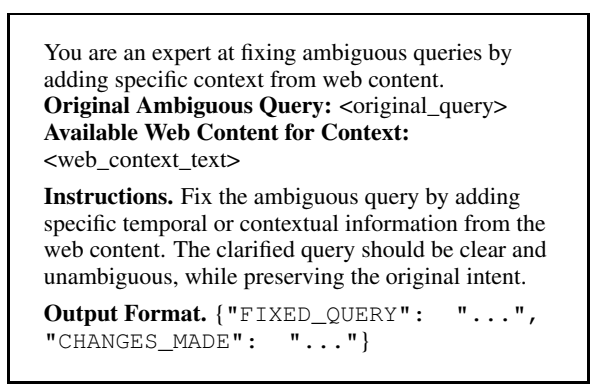

\centering
\begin{tcolorbox}[promptbox,title=]
\small
\RaggedRight
You are an expert at fixing ambiguous queries by adding specific context from web content.

\textbf{Original Ambiguous Query:} \textless original\_query\textgreater  

\textbf{Available Web Content for Context:} \textless web\_context\_text\textgreater

\stitle{Instructions}
Fix the ambiguous query by adding specific temporal or contextual information from the web content. The clarified query should be clear and unambiguous, while preserving the original intent.

\stitle{Output Format}
\{\texttt{"FIXED\_QUERY": "...", "CHANGES\_MADE": "..."}\}
\end{tcolorbox}
\captionsetup{font=small}
\caption{Prompt template for fixing ambiguous queries using contextual information.}
\label{fig:fix_ambiguity}
\end{figure}

\begin{figure}[h]
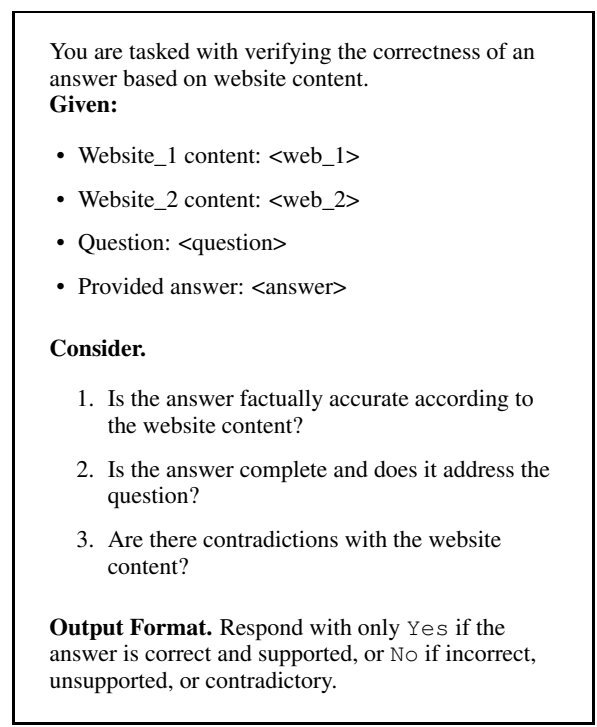

\centering
\begin{tcolorbox}[promptbox,title=]
\small
\RaggedRight
You are tasked with verifying the correctness of an answer based on website content.

\textbf{Given:}
\begin{itemize}[leftmargin=1.2em,itemsep=2pt]
  \item Website\_1 content: \textless web\_1\textgreater
  \item Website\_2 content: \textless web\_2\textgreater
  \item Question: \textless question\textgreater
  \item Provided answer: \textless answer\textgreater
\end{itemize}

\stitle{Consider}
\begin{enumerate}[itemsep=2pt]
  \item Is the answer factually accurate according to the website content?
  \item Is the answer complete and does it address the question?
  \item Are there contradictions with the website content?
\end{enumerate}

\stitle{Output Format}
Respond with only \texttt{Yes} if the answer is correct and supported, or \texttt{No} if incorrect, unsupported, or contradictory.
\end{tcolorbox}
\captionsetup{font=small}
\caption{Prompt template for verifying correctness of answers against website content.}
\label{fig:check_correctness}
\end{figure}

%% file: resources/performance_website.tex
\begin{figure*}[h]
    \centering
    \footnotesize
    \includegraphics[width=1\linewidth]{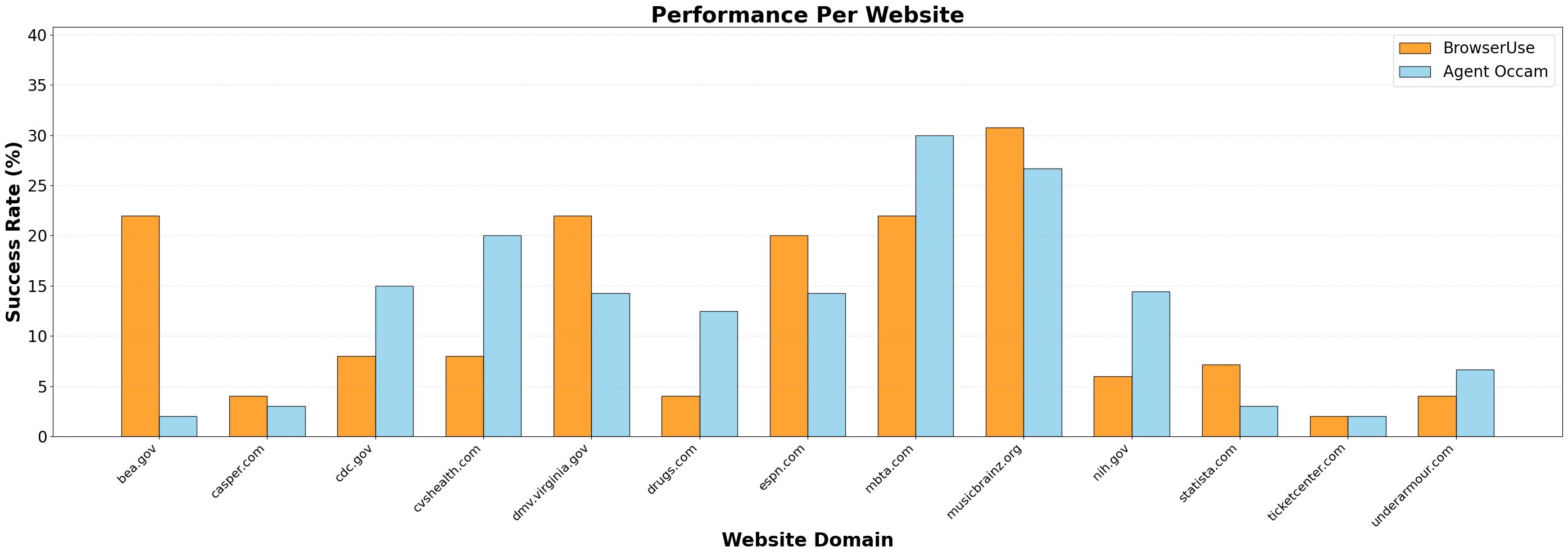}
    \caption{Performance comparison of Browser Use and AgentOccam across representative websites. Each bar shows the average success rate (\%) for completing web-based tasks within a given domain. }
    \label{fig:performance_per_website}
\end{figure*}

%% file: resources/multilingual_task_fig.tex
\begin{figure*}[h]
    \centering
    \footnotesize
    \includegraphics[width=1\linewidth]{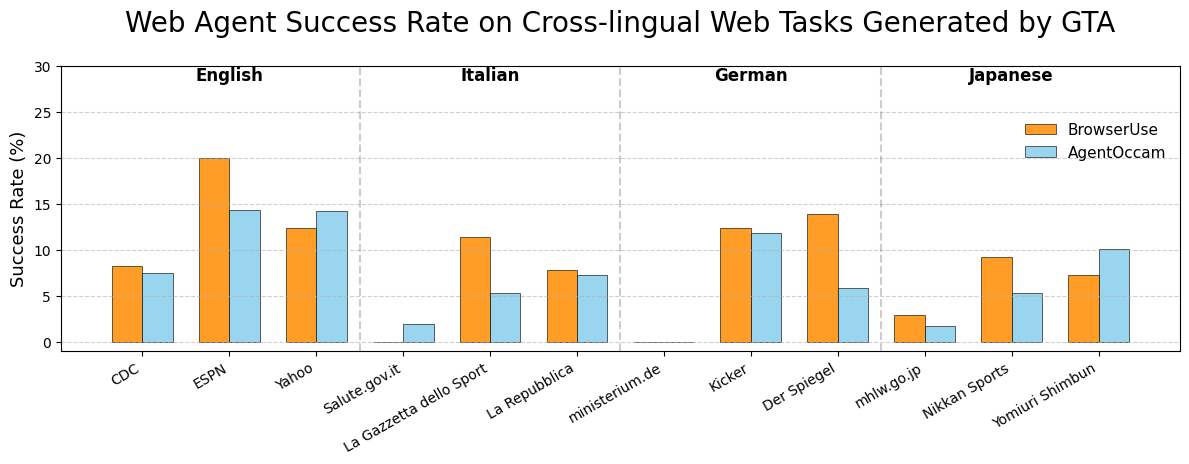}
    \caption{Cross-lingual performance of BrowserUse and AgentOccam performances on web tasks generated by GTA. Each group of bars represents websites in English, Italian, German, and Japanese domains. The overall decline in success rate across languages highlights the growing challenge of multilingual generalization for web agents, especially when adapting to heterogeneous site structures and contents.}
    \label{fig:multilingual}
\end{figure*}

%% file: resources/annotation_inter.tex
\begin{figure*}[h]
    \centering
    \footnotesize
    \includegraphics[width=1\linewidth]{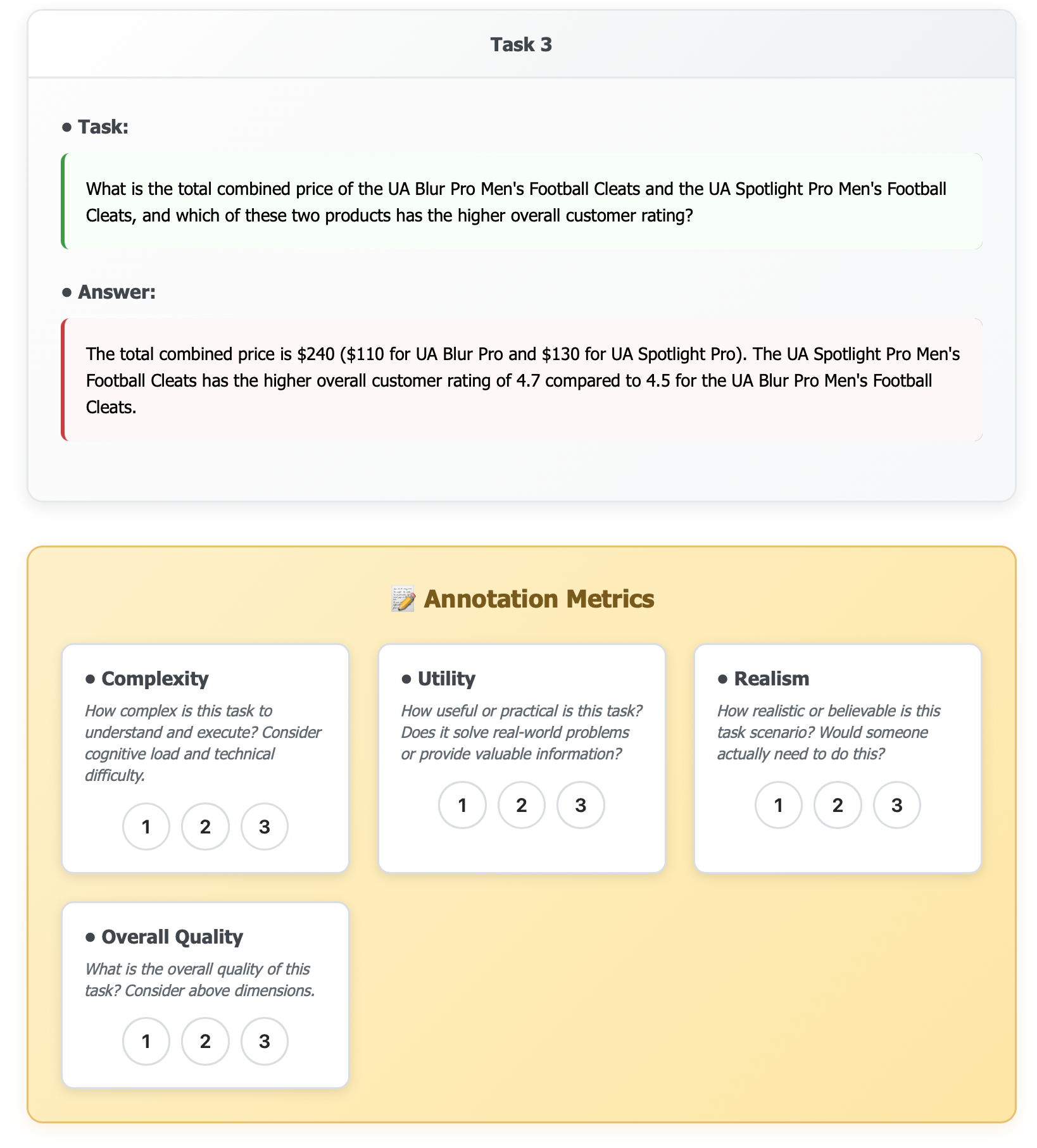}
    \caption{Example annotation interface}
    \label{fig: annotation_interface}
\end{figure*}

%% file: iclr2025.bib
@inproceedings{huang-etal-2025-r2d2,
    title = "{R}2{D}2: Remembering, Replaying and Dynamic Decision Making with a Reflective Agentic Memory",
    author = "Huang, Tenghao  and
      Basu, Kinjal  and
      Abdelaziz, Ibrahim  and
      Kapanipathi, Pavan  and
      May, Jonathan  and
      Chen, Muhao",
    editor = "Che, Wanxiang  and
      Nabende, Joyce  and
      Shutova, Ekaterina  and
      Pilehvar, Mohammad Taher",
    booktitle = "Proceedings of the 63rd Annual Meeting of the Association for Computational Linguistics (Volume 1: Long Papers)",
    month = jul,
    year = "2025",
    address = "Vienna, Austria",
    publisher = "Association for Computational Linguistics",
    url = "https://aclanthology.org/2025.acl-long.1464/",
    doi = "10.18653/v1/2025.acl-long.1464",
    pages = "30318--30330",
    ISBN = "979-8-89176-251-0"
}

@misc{hsu2025webdsendtoendbenchmarkwebbased,
      title={WebDS: An End-to-End Benchmark for Web-based Data Science}, 
      author={Ethan Hsu and Hong Meng Yam and Ines Bouissou and Aaron Murali John and Raj Thota and Josh Koe and Vivek Sarath Putta and G K Dharesan and Alexander Spangher and Shikhar Murty and Tenghao Huang and Christopher D. Manning},
      year={2025},
      eprint={2508.01222},
      archivePrefix={arXiv},
      primaryClass={cs.CL},
      url={https://arxiv.org/abs/2508.01222}, 
}

@article{xu2024agenttrek,
  author    = {Yiheng Xu and Dunjie Lu and Zhennan Shen and Junli Wang and Zekun Wang and Yuchen Mao and Caiming Xiong and Tao Yu},
  title     = {AgentTrek: Agent Trajectory Synthesis via Guiding Replay with Web Tutorials},
  year={2024},
  eprint={2412.09605},
  archivePrefix={arXiv},
  primaryClass={cs.CL},
  url={https://arxiv.org/abs/2412.09605}
}

@inproceedings{
yangagentoccam,
title={AgentOccam: A Simple Yet Strong Baseline for {LLM}-Based Web Agents},
author={Ke Yang and Yao Liu and Sapana Chaudhary and Rasool Fakoor and Pratik Chaudhari and George Karypis and Huzefa Rangwala},
booktitle={The Thirteenth International Conference on Learning Representations},
year={2025},
url={https://openreview.net/forum?id=oWdzUpOlkX}
}

@inproceedings{Murty2025NNetNav,
  author= {Shikhar Murty and Hao Zhu and Dzmitry Bahdanau and Christopher Manning},
  title={NNetNav: Unsupervised Learning of Browser Agents Through Environment Interaction in the Wild},
  journal={arXiv preprint arXiv:2410.02907},
  year={2025}
}

@InProceedings{pmlr-v70-shi17a,
  title = 	 {World of Bits: An Open-Domain Platform for Web-Based Agents},
  author =       {Tianlin Shi and Andrej Karpathy and Linxi Fan and Jonathan Hernandez and Percy Liang},
  booktitle = 	 {Proceedings of the 34th International Conference on Machine Learning},
  pages = 	 {3135--3144},
  year = 	 {2017},
  editor = 	 {Precup, Doina and Teh, Yee Whye},
  volume = 	 {70},
  series = 	 {Proceedings of Machine Learning Research},
  month = 	 {06--11 Aug},
  publisher =    {PMLR},
  url = 	 {https://proceedings.mlr.press/v70/shi17a.html}
}

@inproceedings{liu2018reinforcement,
  title={Reinforcement Learning on Web Interfaces using Workflow-Guided Exploration},
  author={Liu, Evan Zheran and Guu, Kelvin and Pasupat, Panupong and Shi, Tianlin and Liang, Percy},
  booktitle={International Conference on Learning Representations},
  year={2018}
}

@inproceedings{zhouwebarena,
  title={WebArena: A Realistic Web Environment for Building Autonomous Agents},
  author={Zhou, Shuyan and Xu, Frank F and Zhu, Hao and Zhou, Xuhui and Lo, Robert and Sridhar, Abishek and Cheng, Xianyi and Ou, Tianyue and Bisk, Yonatan and Fried, Daniel and others},
year={2024},
  booktitle={The Twelfth International Conference on Learning Representations}

}

@article{Wang2023VoyagerAO,
  title={Voyager: An Open-Ended Embodied Agent with Large Language Models},
  author={Guanzhi Wang and Yuqi Xie and Yunfan Jiang and Ajay Mandlekar and Chaowei Xiao and Yuke Zhu and Linxi (Jim) Fan and Anima Anandkumar},
  journal={Trans. Mach. Learn. Res.},
  year={2023},
  volume={2024},
  url={https://api.semanticscholar.org/CorpusID:258887849}
}

@inproceedings{
sodhi2024step,
title={SteP: Stacked {LLM} Policies for Web Actions},
author={Paloma Sodhi and S.R.K Branavan and Yoav Artzi and Ryan McDonald},
booktitle={First Conference on Language Modeling},
year={2024},
url={https://openreview.net/forum?id=5fg0VtRxgi}
}

@article{deng2023mind2web,
  title={Mind2web: Towards a generalist agent for the web},
  author={Deng, Xiang and Gu, Yu and Zheng, Boyuan and Chen, Shijie and Stevens, Sam and Wang, Boshi and Sun, Huan and Su, Yu},
  journal={Advances in Neural Information Processing Systems},
  volume={36},
  pages={28091--28114},
  year={2023}
}

@article{wu2025webwalker,
  title={Webwalker: Benchmarking llms in web traversal},
  author={Wu, Jialong and Yin, Wenbiao and Jiang, Yong and Wang, Zhenglin and Xi, Zekun and Fang, Runnan and Zhang, Linhai and He, Yulan and Zhou, Deyu and Xie, Pengjun and others},
  journal={arXiv preprint arXiv:2501.07572},
  year={2025}
}

@inproceedings{spangher-etal-2025-creative,
    title = "Creative Planning with Language Models: Practice, Evaluation and Applications",
    author = "Spangher, Alexander  and
      Huang, Tenghao  and
      Laban, Philippe  and
      Peng, Nanyun",
    editor = "Lomeli, Maria  and
      Swayamdipta, Swabha  and
      Zhang, Rui",
    booktitle = "Proceedings of the 2025 Annual Conference of the Nations of the Americas Chapter of the Association for Computational Linguistics: Human Language Technologies (Volume 5: Tutorial Abstracts)",
    month = may,
    year = "2025",
    address = "Albuquerque, New Mexico",
    publisher = "Association for Computational Linguistics",
    url = "https://aclanthology.org/2025.naacl-tutorial.1/",
    doi = "10.18653/v1/2025.naacl-tutorial.1",
    pages = "1--9",
    ISBN = "979-8-89176-193-3"
}

@article{huang2025teaching,
  title={Teaching Language Models To Gather Information Proactively},
  author={Huang, Tenghao and Chen, Sihao and Chen, Muhao and May, Jonathan and Yang, Longqi and Wan, Mengting and Zhou, Pei},
  journal={Findings of the Association for Computational Linguistics: EMNLP 2025},
  pages={15588--15599},
  year={2025}
}

@inproceedings{10.1145/3711896.3737384,
author = {Huang, Tenghao and Lee, Dong Hee and Sweeney, John and Shi, Jiatong and Steliotes, Emily and Lange, Matthew and May, Jonathan and Chen, Muhao},
title = {FoodPuzzle: Toward Developing Large Language Model Agents as Autonomous Flavor Scientists},
year = {2025},
isbn = {9798400714542},
publisher = {Association for Computing Machinery},
address = {New York, NY, USA},
url = {https://doi.org/10.1145/3711896.3737384},
doi = {10.1145/3711896.3737384},
booktitle = {Proceedings of the 31st ACM SIGKDD Conference on Knowledge Discovery and Data Mining V.2},
pages = {5493–5504},
numpages = {12},
location = {Toronto ON, Canada},
series = {KDD '25}
}

@article{wei2025browsecomp,
  title={Browsecomp: A simple yet challenging benchmark for browsing agents},
  author={Wei, Jason and Sun, Zhiqing and Papay, Spencer and McKinney, Scott and Han, Jeffrey and Fulford, Isa and Chung, Hyung Won and Passos, Alex Tachard and Fedus, William and Glaese, Amelia},
  journal={arXiv preprint arXiv:2504.12516},
  year={2025}
}

@article{ishmam2026timewarp,
  title={TimeWarp: Evaluating Web Agents by Revisiting the Past},
  author={Ishmam, Md Farhan and Marino, Kenneth},
  journal={arXiv preprint arXiv:2603.04949},
  year={2026}
}
